\documentclass{article}

\usepackage[preprint, nonatbib]{neurips_2019}

\usepackage[utf8]{inputenc} % allow utf-8 input
\usepackage[T1]{fontenc}    % use 8-bit T1 fonts
\usepackage{hyperref}       % hyperlinks
\usepackage{url}            % simple URL typesetting
\usepackage{booktabs}       % professional-quality tables
\usepackage{amsfonts}       % blackboard math symbols
\usepackage{nicefrac}       % compact symbols for 1/2, etc.
\usepackage{microtype}      % microtypography
\usepackage{times}
\usepackage{latexsym}
\usepackage{url}
\usepackage{color}

\usepackage{graphicx}
\usepackage{amsmath}

\newcommand{\trm}[1]{\textit{{#1}}}
\newcommand{\hs}[1]{\textcolor{blue}{Haitian: #1}} % haitian's comment

\newtheorem{claim}{Claim}

\newcommand{\aq}{q}
\newcommand{\aA}{A}
\newcommand{\aex}{(\aq,\aA)}
\newcommand{\arel}{r}
\newcommand{\aent}{x}
\newcommand{\srel}{R}
\newcommand{\sent}{X}
\newcommand{\avrel}{R}
\newcommand{\alatent}{z}
\newcommand{\aspan}{s}

\newcommand{\followop}{relation-set following}
\newcommand{\follow}[2]{\textit{follow}({#1},{#2})}
\newcommand{\ty}{\textit{type}}

\newcommand{\aty}{\tau}
\newcommand{\aR}{\mathbb{R}}
\newcommand{\vek}[1]{\textbf{#1}}
\newcommand{\aind}{index}
\newcommand{\subj}{{\it subj}}
\newcommand{\obj}{{\it obj}}
\newcommand{\rel}{{\it rel}}

\newcommand{\setw}[2]{\omega|\![{#1} \in {#2}]\!|}
\newcommand{\kdelta}[2]{\delta|\![{#1}={#2}]\!|}

\newcommand{\supp}{\textit{support}}

\title{Differentiable Representations \\
For Multihop Inference Rules}

\author{%
  William W. Cohen \\ Google Research \\ \texttt{wcohen@google.com} \\
  \And
  Haitian Sun \\ Google Research \\ \texttt{haitiansun@google.com} \\
  \And
  R. Alex Hofer \\ Google \\ \texttt{rofer@google.com} \\
  \And
  Matthew Siegler \\ Google Research \\
  \texttt{msiegler@google.com} \\
}

\begin{document}

\maketitle

\begin{abstract}
We present efficient differentiable
implementations of second-order multi-hop reasoning using a large
symbolic knowledge base (KB).  
We introduce a new operation which can be used to compositionally
construct second-order multi-hop templates in a neural model, and
evaluate a number of alternative implementations,
with different time and memory trade offs.
These techniques scale to KBs with millions of entities and tens
of millions of triples, and lead to simple models with competitive
performance on several learning tasks requiring multi-hop reasoning.
\end{abstract}

\vspace{-0.05in}

\section{Introduction}

\vspace{-0.05in}

\textbf{Templates for multi-hop reasoning.}  For certain applications it is useful for a neural model to be able to encode multi-step accesses to a symbolic KB.  For example, for the task of learning to answer natural-language questions with a KB, a question like $\aq$ = ``who has directed a movie written by
Christopher Nolan?'' might be answered with the set 
\( \{ \aent' : \exists
    \aent,\alatent \mbox{~so that~}
   \textit{referent}(\textit{``Christopher Nolan''},\aent) \wedge
   \textit{writer\_of}(\aent,\alatent) \wedge
   \textit{director\_of}(\aent',\alatent) \} 
\). Computing this set
requires \textit{multi-hop reasoning}, i.e., 
multiple steps of KB access: in this case, finding the referent
of the name ``Christopher Nolan'', then finding the movies written by that
person, and then finding the directors of those movies.
Constructing a logical interpretation like this from text  is called
\trm{semantic parsing}. 

Since the chains of reasoning
in parses are usually short,  
semantic parsing often can be reduced to finding an appropriate template and
instantiating it. For instance, queries like $\aq$ above could be modeled with the template
$$
      \{ \aent' : \exists \aent,\alatent,\aent',\aspan,\avrel_1,\avrel_2 \textrm{~so that~} 
         \textit{subspan}(\aq, \aspan) \wedge
         \textit{referent}(\aspan, \aent) \wedge \avrel_1(\aent, \alatent) \wedge \avrel_2(\alatent, \aent') \}
$$
where $\aspan$ is a variable ranging over subspans of $\aq$; $\aent$, $\aent'$, and $\alatent$
range over entities, and $\avrel_1$ and $\avrel_2$ range over relations. Logical
expressions with variables ranging over relations are usually
called \textit{second-order expressions} or \trm{templates}.  

\textbf{Differentiable multi-hop templates.}
Learning a template-based semantic parser requires learning to select and instantiate the template, and instantiation requires learning a function for each variable in the template---e.g., a system might let $\aspan=f_\aspan(\aq)$,  $\avrel_1=f_{\avrel_1}(\aq)$, and
$\avrel_2=f_{\avrel_2}(\aq)$, where $f_\aspan$, $f_{\avrel_1}$ and $f_{\avrel_2}$ might be, for instance, a softmax functions applied to linear transformations of a biLSTM-based encoding of $\aq$.  Training these extractors requires data 
in which each $\aq$ is labeled with the value of each template variable---e.g., here labels for $\aq$ would specify that $f_{\avrel_1}(\aq)$ should be the relation
\textit{writer\_of}, etc. It would be preferable to be able to train  extractors
end-to-end, using as training examples pairs $\aex$, where $\aq$ is a
document and $\aA$ is the result of evaluating the desired template---e.g., here $\aA$ would be all KB entities
representing people that directed Nolan's movies. However,
while there is a literature on training semantic parsers end-to-end (e.g., \cite{berant2013semantic,yih2015semantic})
learning in this setting with modern gradient approaches requires evaluating the second-order template with only
differentiable operations. Further, these evaluations must be done very
efficiently over large KBs, since they are performed at training time; hence, to our knowledge, this approach has been used only when question-related knowledge is limited \cite{zhong2017seq2sql,DBLP:journals/corr/abs-1808-09942,graftnet}.

This paper describes a framework for efficiently evaluating multi-hop templates differentiably, enabling, for the first time, deep learning in end-to-end settings that require reasoning over large symbolic KBs.
This is based on a primitive operation called \trm{\followop{}}, which finds all KBs entities related to any member of an input set $X$ via any relation in a set $R$.  By nesting this operation, multi-step reasoning templates can be formed.  We present results for 
semantic parsing using second-order templates and KB completion, which can also be formulated as a multi-hop second-order reasoning task.

\vspace{-0.06in}

\section{Differentiable Templates for Multi-Hop Reasoning}

\vspace{-0.04in}

\textbf{Preliminaries: KBs, entities, and relations.}  A KB consists
of \trm{entities} and \trm{relations}.  We use $\aent$ to denote an
entity and $\arel$ to denote a relation.  A relation is a set of
entity pairs, and represents a relationship between entities: for
instance, if $\aent$ represents
``Christopher Nolan'' and $\aent'$ represents ``Inception'' then $(\aent,\aent')$ would be an member of the
relation \textit{writer\_of}.  If $(\aent,\aent')\in\arel$ we say that
\trm{$\arel(\aent,\aent')$ is an assertion (in the KB)}.

We assume each entity $\aent$ has a \trm{type}, written $\ty(\aent)$, and
let $N_\aty$ denote the number of entities of type $\aty$.  Each
entity $\aent$ in type $\aty$ also has a unique index
$\aind_\aty(\aent)$, which is an integer between $1$ and $N_{\aty}$.
We write $\aent_{\aty,i}$ for the entity that has index $i$ in type
$\aty$, or $\aent_i$ if the type is clear from context.  
%The set of all entities in a KB of type $\aty$
%is written $_\aty$.

Every relation $\arel$ has a \trm{subject type} $\aty_\subj$ and an
\trm{object type} $\aty_\obj$, which constrain the types of $\aent$
and $\aent'$ for any pair $(\aent,\aent') \in \arel$.  Hence $\arel$
can be encoded as a subset of \( \{1,\ldots,N_{\aty_\subj}\} \times
\{1,\ldots,N_{\aty_\obj}\} \).  Relations with the same subject and
object types are called \textit{type-compatible}.
Finally a KB consists of a set of types, a set of typed relations, and
a set of typed entities.  

\textbf{Weighted sets and their encodings.} Our differentiable
operations are based on \trm{weighted sets}, where each element
$\aent$ of weighted set $\sent$ is associated with a non-negative real
number, written $\setw{\aent}{\sent}$.  It is convenient to define
$\setw{\aent}{\sent}\equiv{}0$ for all $\aent\not\in\sent$.
Conceptually, a weight less than 1 for element $\aent$ is a confidence
that the set contains $\aent$, and weights more than 1 make $\sent$ a
multiset.  If all elements of $\sent$ have weights 1, we say $\aent$
is a \trm{hard set}. Weighted sets $\sent$ are also typed, and if
$\ty(\sent)=\aty$ then $\sent$ is constrained to contain only entities
of type $\aty$; hence $\sent$ can be encoded as a subset of
$\{1,\ldots,N_{\ty(\sent)}\}$.

Weighted sets will be used as potential assignments to the variables
in second-order templates.  To support second-order reasoning, every
relation $\arel$ in a KB is also associated with an entity
$\aent_\arel$, and hence, an index and a type, and sets of relations
$\srel$ are allowed if all members are type-compatible.\footnote{To
  avoid complexity in notation we do not distinguish between sets of
  relations and sets of entities that are associated relations.}  For
example $\srel = \{ \textit{writer\_of}, \textit{director\_of} \}$
might be a set of type-compatible relations.  For notational
convenience, we also treat the relations in a KB as weighted sets of
entity pairs, although in the experiments of this paper, all KB
relations are hard sets.

%\subsection{Numerically encoding a KB}

A weighted set $\sent$ of type $\aty$ can be encoded as an
\trm{entity-set vector} $\vek{v}_\sent \in \aR^{N_\aty}$, where the
$i$-th component of $\vek{v}_\sent$ is the weight of the $i$-th entity
of that type in the set $\sent$: e.g., \(
\vek{v}_\sent[\aind_\aty(\aent)] = \setw{\aent}{\sent} \). Notice that
if $\sent$ is a hard entity set, then this will be a $k$-hot vector
for $k=|\sent|$.  The set of indices of $\vek{v}$ with non-zero values
is written $\supp(\vek{v})$, and we also use $\ty(\vek{v})$ to denote
the type $\aty$ of the set encoded by $\vek{v}$.  A relation $\arel$
with subject type $\aty_1$ and object type $\aty_2$ can be encoded as
a \trm{relation matrix} $\vek{M}_\arel \in \aR^{N_{\aty_1} \times
  N_{\aty_2}}$, where the value for $\vek{M}_\arel[i,j]$ is the
weight of the assertion $\arel(\aent_i,\aent_j)$ in the KB, i.e., \(
\vek{M}_\arel[\aind_{\aty_1}(\aent),\aind_{\aty_2}(\aent')] =
\setw{(\aent,\aent')}{\arel} \).

\textbf{Sparse relation matrices.}
For all but the smallest KBs, a
relation matrix must be implemented using a \trm{sparse matrix} data
structure, as explicitly storing the $N_{\aty_1} \times N_{\aty_2}$
values will be very wasteful. For instance, if a KB contains 10,000
movie entities and 100,000 person entities, then a relationship like
\textit{writer\_of} would require storing 1 billion values, while,
since most movies have only a few writers, only a few tens of
thousands of \textit{writer\_of} facts will be in the KB.  
One possible sparse matrix data structure for a relation matrix is a
\trm{sparse coordinate pair (COO)} structure, which consists of a
$N_\arel \times 2$ matrix $\vek{Ind}_\arel$ containing pairs of entity
indices, and a parallel vector $\vek{w}_\arel \in \aR^{N_\arel}$
containing the weights of each entity pair in $M_\arel$.  In this
encoding, if $(i,j)$ is row $k$ of $\vek{Ind}$, then
$\vek{M}_\arel[i,j] = \vek{w}_\arel[k]$, and if $(i,j)$ does not
appear in $\vek{Ind}_\arel$, then $\vek{M}[i,j]$ is zero.  Hence with
a COO encoding, the size needed to encode the relations in the KB is
linear in the number of facts.

Our experiments are performed with Tensorflow \cite{abadi2016tensorflow}, which offers
limited support for sparse matrices.  In particular, Tensorflow supports
sparse COO matrices, but not higher-rank tensors, and supports matrix
multiplication between a sparse matrix and a dense matrix, but not
between two sparse matrices.  %Since sparse matrices are used for
%relation matrices, this means that entity-set vectors must be ordinary
%dense vectors.

\vspace{-0.04in}

\subsection{Reasoning in a KB}

\vspace{-0.03in}

\textbf{Single-hop reasoning and $\srel$-neighbors.}
Relations can also be viewed as sets of labeled edges in a
\trm{knowledge graph}, the vertices of which are entities.  Following
this view, we define the \trm{$\arel$-neighbors of an entity $\aent$}
to be the set of entities $\aent'$ that are connected to $\aent$
by an edge labeled $\arel$, i.e.,
\( \textit{$\arel$-neighbors($\aent$)} \equiv \{ \aent' : (\aent,\aent') \in \arel \}
\).
Extending this to sets, we define
\[ \textit{$\srel$-neighbors($\sent$)} \equiv \{ \aent' : \exists \arel\in\srel, \aent\in\sent \mbox{~so that~} (\aent,\aent') \in \arel \}
\]
Computing the $\srel$-neighbors of an entity is a single-step
reasoning operation: e.g., the answer to the question $q=$``what
movies were written by Christopher Nolan'' is precisely the set
$\srel$-neighbors($\sent$) for $\srel=\{\textit{writer\_of}\}$ and
$\sent=\{\textit{Christopher\_Nolan}\}$. Multi-hop reasoning
operations require nested $\srel$-neighborhoods, e.g.  if
$\srel'=\{\textit{director\_of}\}$ then 
$\srel'$-{neighbors}($\srel$-{neighbors}$(\sent))$ is all
directors of movies written by Christopher Nolan.

\textbf{Encoding single-hop reasoning.} We would like to ``soften''
the $\srel$-neighbors, and also translate it into differentiable
operations that can be performed on encodings of $\sent$ and $\srel$.
Let $\vek{v}_\sent$ encode a weighted set of entities $\sent$, and let
$\vek{v}_\srel$ encode a weighted set of relations of type $\aty$.  We
first define $\vek{M}_\srel$ to be a weighted mixture of the relation
matrices for all relations of type $\aty$, i.e., \( \vek{M}_\srel
\equiv (\sum_{k=1}^{N_\aty} \vek{v}_\srel[k] \cdot \vek{M}_{r_k}) \).
%Note that this is well-defined, since the matrices being summed are
%for type-compatible relations, which therefore have the same
%shapes. 
We then define the \trm{\followop{} operation for
  $\vek{v}_\sent$ and $\vek{v}_\srel$} as:
\begin{equation}
 \label{eq:followdef}
 %x
    \follow{\vek{v}_\sent}{\vek{v}_\srel}   \equiv
     \vek{v}_\sent \vek{M}_\srel 
     = \vek{v}_\sent ( \sum_{k=1}^{N_\aty} \vek{v}_\srel[k] \cdot \vek{M}_k )
\end{equation}

\vspace{-0.1in}

The numerical \followop{} operation of Eq~\ref{eq:followdef} corresponds
closely to the logical $\srel$-neighborhood operation, as shown by the
claim below.

\vspace{-0.05in}

\begin{claim} The support of $\follow{\vek{v}_\sent}{\vek{v}_\srel}$ is
exactly the set of $\srel$-neighbors($\sent$).
\end{claim}

\vspace{-0.05in}

\iffalse
To see this, consider first a matrix $\vek{M}_\arel$ encoding a single
binary relation $\arel$, and consider the vector $\vek{v}_{\sent'} =
\vek{v}_\sent \vek{M}_\arel$.  As weighted sets, $\sent$ and $r$
have non-negative entries, so clearly for all $i$,
$$ \vek{v}_{\sent'}[j] \not= 0 \mbox{~~iff~~} 
 \exists j : \vek{M}_\arel[i,j] \not= 0 \wedge \vek{v}_\sent[i] \not= 0 
 \mbox{~~iff~~} \exists \aent_i\in \sent \mbox{~so that~} (\aent_i,\aent_j) \in \arel
$$ and so if $\vek{v}_\arel$ is a one-hot vector for the set
 $\{\arel\}$, then $\supp(\follow{\vek{v}_\sent}{\vek{v}_\arel})$ is
 exactly the set $\arel$-neighbors($\sent$).  Finally note that the
 mixture $\vek{M}_\srel$ has the property that
 $\vek{M}_\srel[i(e_1),i(e_2)] > 0$ exactly when $e_1$ is related to
 $e_2$ by some relation $\arel \in \srel$.
 \fi

To better understand this claim, let $\vek{z}=\follow{\vek{v}_\sent}{\vek{v}_\srel}$.  The claim
states $\vek{z}$ can approximate the $\srel$ neighborhood of any
hard sets $\srel,\sent$ by setting to zero the appropriate components
of $\vek{v}_\sent$ and $\vek{v}_\srel$.  It is also clear that
$\vek{z}[j]$ decreases when one decreases the weights in
$\vek{v}_\arel$ of the relations that link $\aent_j$ to entities in
$\sent$, and likewise, $\vek{z}[j]$ decreases if one
decreases the weights of the entities in $\sent$ that are linked to
$\aent_j$ via relations in $\srel$, so there is a
smooth, differentiable path to reach this approximation.

The utility of approximating the single-step inference step of
computing $\arel$ neighborhoods with the \followop{} is further
supported by the experimental results below.
\iftrue
We will also see below how to produce more complex templates by combining multiple \followop{} operations.

\else
\subsection{Multi-step reasoning templates}

It is quite simple to combine these operations to perform more complex
reasoning tasks.  A two-hop question like the one in the introduction
could be implemented with a model that computes 
$$ \vek{\arel}_1 = f_1(q);~~ \vek{\arel}_2 = f_2(q);~~ \vek{\aent} = f_\aent(q);~~
   \hat{\vek{a}} = \follow{\follow{\vek{\aent}}{\vek{\arel}_1}}{\vek{\arel}_2}
$$ 
If one would like to be able answer a mix of two-hop and one-hop
questions, one might introduce a new variable $p=f_p(q)$ which
estimates the probability of a two-hop question, and replace the last
step with \hs{change to bold symbols}
$$ \hat{\vek{a}} = p \cdot \follow{\follow{\vek{\aent}}{\vek{\arel}_1}}{\vek{\arel}_2} + (1-p)\cdot \follow{\vek{\aent}}{\vek{\arel}_1}
$$ Extending the model to evaluate sets like ``movies written and
produced by Christopher Nolan'' would require adding a new template to
the mixture of the form
$\follow{\vek{\aent}}{\vek{\arel}_1}\odot\follow{\vek{\aent}}{\vek{\arel}_2}$. Experiments
below show that simple mixtures of this sort suffice for several
widely-used benchmarks---perhaps a consequence of the fact that
extremely complex linguistic constructions are uncommon.  We will also
present an encoder-decoder approach that constructs reasoning chains
dynamically, and evaluate the approach on a synthetic dataset that
requires very long chains of reasoning.
\fi

\begin{table}[t]
    \centering
    \begin{tabular}{c|c|c|ccc}
    \hline
         Strategy &  Batch? & Space complexity & \multicolumn{3}{c}{\# Operations}\\
         & & & sp-dense & dense  & sparse\\
         & & & matmul   & + or $\odot$    & + \\

         \hline
         naive       & no  & $O(N_T + N_E + N_R)$     & 1 & 0 & $N_R$ \\
         late mixing & yes & $O(N_T + b N_E + b N_R)$ & $N_R$ & $N_R$ & 0 \\
         reified KB  & yes & $O(b N_T + b N_E)$       & 3     & 1     & 0 \\
         \hline
    \end{tabular}\\
    ~\\
    \caption{Summary of  implementations of \followop{}, where $N_T$ is the number of KB triples, $N_E$ the number of entities, $N_R$ the number of relations.    }
    \label{tab:implementations}
    \vspace{-0.25in}
\end{table}

\textbf{Implementations of \followop.}
Ignoring types for the moment, suppose the KB contains $N_R$
relations, $N_E$ entities, and $N_T$ triples.  Typically $N_R < N_E <
N_T \ll N_E^2$.  As noted above, we must implement each
$\vek{M}_\arel$ as a sparse matrix, so collectively these matrices
require space $O(N_T)$.  Each triple appears in only one relation, so
$\vek{M}_\srel$ is also size $O(N_T)$.  Since sparse-sparse matrix
multiplication is not supported in Tensorflow we must implement
$\vek{x}\vek{M}_\srel$ using dense-sparse multiplication\footnote{In
  fact Tensorflow requires the \emph{left} multiplicand to be sparse,
  so really we must use dense-sparse matrix multiplication and compute
  $(\vek{M}_k^T \vek{x}^T)^T$.}, so $\vek{x}$ must be a
dense vector of size $O(N_E)$, as is the output of \followop{}. So
the space complexity of $\follow{\vek{x}}{\vek{r}}$ is $O(N_T + N_E +
N_R)$, if implemented as suggested by Eq~\ref{eq:followdef}; see Table~\ref{tab:implementations}.
%The
%implementation also requires $N_R$ sparse matrix summations, which
%(since they require deduping sets of indices) are relatively expensive
%to perform on a GPU.
The major problem with this implementation is that, in the absence of
general sparse tensor contractions, it is difficult to adapt to
mini-batches, which usually make inference on GPUs is much faster. We
thus call this implementation \trm{naive mixing} and, in this paper,
only use it without minibatches.  

We consider next a setting in which $\vek{x}$ and $\vek{r}$ are
replaced with matrices $\vek{X}$ and $\vek{R}$ with the minibatch size
$b$.  An alternative strategy is based on the observation that
\followop{} for a single relation can be implemented as
$\vek{x}\vek{M}_\arel$, which can be trivially extended to a minibatch
as $\vek{X}\vek{M}_\arel$.  The \trm{late mixing} strategy mixes the
output of many single-relation \followop{} steps, rather than mixing
the KB:
\begin{equation}
 \label{eq:latemix}
    \follow{\vek{X}}{\vek{R}} = 
       \sum_{k=1}^{N_\aty} (\vek{R}[:,k] \cdot \vek{X} \vek{M}_k)
\end{equation}
where the $k$-th column of $\vek{R}$ is ``broadcast'' to each element of the matrix $\vek{X} \vek{M}_k$.  While there are $N_R$ matrices $\vek{X} \vek{M}_k$, each of size $O(b N_E$), they need not all be stored at once to be mixed, so the space complexity becomes $O(b N_E + b N_R + N_T)$.  However, we now need to sum up $N_R$ dense matrices. 

An alternative to late mixing is to represent the KB with number of
matrices.  Each KB assertion $\arel(e_1,e_2)$ can be represented as a
tuple $(i, j, k, w)$ where $i,j,k$ are the indices of $e_1,e_2$, and $\arel$,
and $w$ is the confidence associated the triple.  There are $N_T$ such
triples, so for $\ell=1,\ldots,N_T$ let 
\( (i_\ell, j_\ell, k_\ell,
w_\ell) \)  denote the $\ell$-th triple.  Let $\kdelta{a}{b}$ denote 1 if
$a=b$ and 0 otherwise. We now define these sparse matrices, which
collectively define the \trm{reified KB}:\footnote{Reification in
  logic is related to reflection in programming languages.}
$$
\vek{M}_\subj[\ell,m] \equiv \kdelta{m}{i_\ell} ~;~~
\vek{M}_\obj[\ell,m] \equiv \kdelta{m}{j_\ell}  ~;~~ 
\vek{M}_\rel[\ell,m] \equiv w_\ell\cdot \kdelta{m}{k_\ell}
%begin{itemize}
%item $\vek{M}_\subj[\ell,m]=1$ if $m=i_\ell$, and %$\vek{M}_\subj[\ell,m]=0$ otherwise. 
%\item $\vek{M}_\obj[\ell,m]=1$ if $m=j_\ell$, and %$\vek{M}_\obj[\ell,m]=0$ otherwise. 
%\item $\vek{M}_\rel[\ell,m]=w_\ell$ if $m=k_\ell$, and %$\vek{M}_\rel[\ell,m]=0$ otherwise. 
%\end{itemize}
$$ Conceptually, $\vek{M}_\subj$ maps the index $\ell$ of the
$\ell$-th triple to its subject entity; $\vek{M}_\obj$ maps $\ell$ to
the object entity; and $\vek{M}_\rel$ maps $\ell$ to the relation of
the $\ell$-th triple, and incidentally encodes the triple confidence
$w_\ell$.  We can now implement the \followop{} as below,
where $\odot$ is Hadamard product:
\begin{equation}
 \label{eq:rjoin}
    \follow{\vek{X}}{\vek{R}} = 
    ( \vek{X}\vek{M}^T_\subj \odot \vek{R}\vek{M}^T_\rel) \vek{M}_\obj
\end{equation}
For a single $\vek{x},\vek{r}$ notice that $\vek{x}\vek{M}^T_\subj$
are triples with entity in $\vek{x}$ as their subject,
$\vek{r}\vek{M}^T_\rel$ are the triples with a relation in $\vek{r}$,
and the Hadamard product is the intersection of these.  The final
multiplication by \( \vek{M}_\obj \) finds the object entities of the
triples in the intersection.  These operations naturally extend to
minibatches, as given in Eq~\ref{eq:rjoin}.  The reified KB has size
$O(N_T)$, the sets of triples that are intersected have size \( O(b
N_T) \), and the final result is size $O(b N_E)$, giving a final size
of $O(b N_T + b N_E)$, with no dependence on $N_R$.

These alternative implementations are summarized in
Table~\ref{tab:implementations}.  
Note that no strategy dominates, but
the analysis suggests that the reified KB is preferable if there are
many relations, while the late mixing strategy may be
preferable if $N_R$ is small or if the KB is large. 
%\footnote{
%It should be noted that the asymptotic analysis above conceals a
%meaningful difference in storage when $N_T$ is large: while late
%mixing requires 3 words per KB triple, independent of batch size, the
%matrices of a reified KB require 9 words per triple, plus $2b N_T$
%words to hold the computation: hence even for $b=1$ the reified KB
%requires almost four times as much space as early mixing.}

\textbf{Distributed computation for large KBs.}
Since GPU memory is limited, we also considered distributed computation of the \followop{}. In general matrix multiplication $\vek{X}\vek{M}$ can be decomposed:  $\vek{X}$ can be split into  a ``horizontal stacking'' of $m$ submatrices,
which we write as \( \left[ \vek{X}_1 ; \ldots  ;\vek{X}_m \right] \), and $\vek{M}$ can be similarly partitioned into $m^2$ submatrices, and then we note that
$$ \vek{X} \vek{M} = 
\left[ \vek{X}_1 ; \vek{X}_2; \ldots  ;\vek{X}_m \right] 
\left[ 
    \begin{array}{cccc}
    \vek{M}_{1,1} & \vek{M}_{1,2} & \ldots & \vek{M}_{1,m}\\
    \vdots & \vdots &  & \vdots \\
    \vek{M}_{m,1} & \vek{M}_{m,2} & \ldots & \vek{M}_{m,m}\\
    \end{array} \right] =
    \left[
      (\sum_{i=1}^m \vek{X}_1 \vek{M}_{i,1}); \ldots; 
      (\sum_{i=1}^m \vek{X}_m \vek{M}_{i,m})
    \right]
$$ This can be computed without storing either $\vek{X}$ or $\vek{M}$ on a single machine.  In our implementation of reified KBs, we distribute the matrices that define a reified KB ``horizontally'', so that different triple ids $\ell$ are stored on different GPUs. 

%\vspace{-0.05in}

\section{Experiments}

%\vspace{-0.1in}

\subsection{Scalability} \label{sec:scalability}

%\vspace{-0.05in}

\begin{figure}[t]
\centering
    \includegraphics[width=\textwidth]{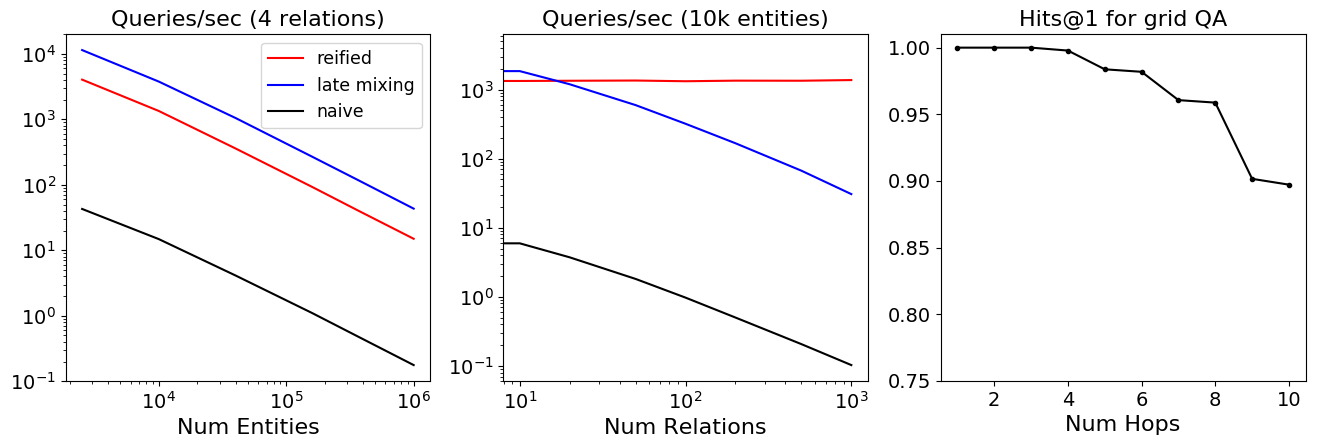}
    \caption{Left and middle: inference time in queries/sec on a synthetic KB.  Right: accuracy on long inference chains for a synthetic QA task, also defined on grids.}
    \label{fig:scaling}
    \vspace{-0.1in}
\end{figure}

Following prior work \cite{cohen2017tensorlog,de2007problog}, we used
a synthetic KB based on an $n$-by-$n$ grid to study scalability of
inference on large KBs. Every grid cell is an entity, related to its
immediate neighbors, via relations \textit{north}, \textit{south},
\textit{east}, and \textit{west}.  The KB for $n$-by-$n$ thus has
$n^2$ entities and around $4n$ triples (since edge cells have fewer
than four neighbors).  We measured the time to compute \(
\follow{\follow{\vek{X}}{\vek{R}}}{\vek{R}} \) for minibatches of
$b=128$ one-hot vectors, and report it as queries per second (qps) on a
single GPU (e.g., qps=1280 would mean a single minibatch requires
100ms).  The matrix $\vek{R}$ weights all relations uniformly, and we
vary the number of relations by inventing $m$ new relation names and
assigning one of the existing grid triples to each new relation.

The results are shown Figure~\ref{fig:scaling} (left and middle), on a
log-log scale because some differences are very large.  With only four
relations (the leftmost plot), late mixing is consistently about 3x
faster than the reified KB method, and about 250x than the naive
approach.  For more than around 20 relations, the reified KB is
faster: it is about 43x faster than late mixing with 1000 relations,
and more than 12,000x faster than the naive approach.  

Although we do not show the results, the method we call ``naive'' here
is much more memory-efficient than using dense relation-matrices. We
do not show results for smaller minibatch sizes, but they are about
40x slower with $b=1$ than with $b=128$ for both reified and late
mixing.

%\vspace{-0.07in}

\subsection{Quality and simplicity of models}

%\vspace{-0.05in}

Below we present results on several datasets using models that make
diverse use of \followop{}.  All experiments in this section are
performed with the reified KB implementation.  Since the main
contribution of this paper is to introduce a generally useful,
scalable, differentiable, second-order reasoning operation, we
introduce models that use \followop{} but are otherwise quite simple,
rather than complex architectures carefully crafted for a particular
task.

% \yy{webquestionsSP}
\subsubsection{Question Answering Experiments}
\textbf{End-to-end QA with a large KB.} WebQuestionsSP
\cite{yih2016value} contains 4737 questions posed in natural
languages, all of which are answerable using Freebase
\cite{bollacker2008freebase}.  Each question $\aq$ is labeled with the
entities $\vek{x}$ that appear in it.  In our experiments we used a
subset of Freebase with 43 million facts and 12 million entities---the
KB includes all facts in Freebase within 2-hops of entities mentioned
in any question, excluding paths through some very common entities.

Freebase contains two types of nodes, one for real-world entities, and
one for \textit{compound value types} (CVTs), which often represent
non-binary relationships or events (e.g., a movie release event, which
includes a movie id, a date, and a place.)  In this dataset, all
questions can be answered with 1- or 2-hop chains, and all 2-hop
reasoning chains pass through a CVT entity.  Hence there are only two
possible templates, which use three kinds of relations: relations
between two ordinary entities, which are used for 1-hop questions;
relations between an ordinary entity and a CVT entity, which are used
for the first step of a 2-hop chain; and relations between a CVT and
an ordinary entity, which are used for the second step of a 2-hop
chain.  Our model begins by deriving from $q$ three relation sets, one
of each kind listed above, and then uniformly mixes together the two
templates and applies a softmax:
\begin{equation*}
\begin{gathered}
\vek{r}_{\textnormal{E} \rightarrow \textnormal{E}} = f_{\textnormal{E} \rightarrow \textnormal{E}}(q); ~~~
\vek{r}_{\textnormal{E} \rightarrow \textnormal{CVT}} = f_{\textnormal{E} \rightarrow \textnormal{CVT}}(q); ~~~
\vek{r}_{\textnormal{CVT} \rightarrow \textnormal{E}} = f_{\textnormal{CVT} \rightarrow \textnormal{E}}(q)
\\
\hat{\vek{a}} = \textit{softmax}(\follow{\follow{\vek{x}}{\vek{r}_{\textnormal{E} \rightarrow \textnormal{CVT}}}}{\vek{r}_{\textnormal{CVT} \rightarrow \textnormal{E}}} + \follow{\vek{x}}{\vek{r}_{\textnormal{E} \rightarrow \textnormal{E}}})
\end{gathered}
\end{equation*}
Note the 2-hop template contains nested \followop{} operations.

In our model, $f_{\textnormal{E} \rightarrow \textnormal{E}}$,
$f_{\textnormal{E} \rightarrow \textnormal{CVT}}$, and
$f_{\textnormal{CVT} \rightarrow \textnormal{E}}$ are each linear
projections of a common encoding for $q$, which is a mean-pooling of
the tokens in $q$ encoded with a pre-trained 128-dimensional word2vec
model \cite{mikolov2013distributed}. We use unregularized cross
entropy loss.  For this large KB, we split the KB across three 12-Gb
GPUs, and used a fourth GPU for the rest of the model.  Performance
(using Hits@1) is shown in Table \ref{tbl:exp_realqa}, in the first
column, as the model RSF (for \followop{}).

%\begin{table}[htp]
%\centering
%\small
%\begin{tabular}{lccc}
%\hline
%& KV-Mem & GRAFT-Net & NQL (ours) \\ \hline
%Hits@1 & 46.7   & 67.8      & 52.7      \\ \hline
%\end{tabular}
%\caption{Experiment results for WebQuestionsSP dataset.\label{tbl:exp_webqsp}}

\begin{table}[t]
\centering
\small
\begin{tabular}[t]{r|c|ccc}
\hline
          & WebQSP      & \multicolumn{3}{|c}{MetaQA} \\
          & 1-2 hops   & 1-hop & 2-hop & 3-hop \\ \hline
KV-Mem*    &  46.7          & 95.8           & 25.1           & 10.1           \\
VRN*       &  ---           & 97.5           & 89.9           & 62.5           \\
GRAFT-Net* & 67.8          & 97.0           & 94.8           & 77.7           \\
RSF (ours)& 52.7           & 96.2           & 81.1           & 72.3           \\
%PullNet* & & & & \\
\hline
\end{tabular}
~~~~~~~~~~\begin{tabular}[t]{r|cc}
\hline
     & \multicolumn{2}{|c}{NELL-995} \\
     & H@1 & H@10 \\
     \hline
%ComplEx* & 61.2 & 82.7 \\
DistMult*& 61.0 & 79.5 \\
ConvE*   & 67.2 & 86.4 \\
MINERVA  & 66.3 & 83.1 \\
RSF (Ours)& 64.1 & 82.4\\
\hline
\end{tabular}\\
~\\
\caption{\textbf{Left:} Hits@1 for the WebQuestionsSP and MetaQA datasets. Results for KV-Mem and VRN on MetaQA are from \cite{zhang2017variational}; results for GRAFT-Net and KV-Mem on WebQSP are from \cite{graftnet}.  Systems marked with a star $*$ construct question-specific autographs, based on the KB, at inference time.  \textbf{Right:} Hits@1 and hits@10 for
two KB completion tasks.  Starred KB completion methods are
transductive, and do not generalize to entities not seen in training.
\label{tbl:exp_realqa}}
\vspace{-0.25in}
\end{table}
 
\textbf{End-to-end QA for longer reasoning chains.} MetaQA \cite{zhang2017variational} consists of 1.2M questions, of which 1/3 are 1-hop, 1/3 are 2-hop, and 1/3 are 3-hop.  The questions 
also are labeled with entities and can be answered 
using the KB provided by the WikiMovies \cite{kvmem} dataset, which contains 43k entities and 186k facts.  Following past work, we train and test our model with 1-hop, 2-hop or 3-hop questions separately.  For each step of inference we construct a relation sets $\vek{r}$.  Let $\vek{x}^0$ be the set of entities associated with $q$. The full model computes is
$$
\textnormal{for $t=1,2,3$:}~~~
\vek{r}^t = f^{t}(q); ~~~
\vek{x}^{t} = \follow{\vek{x}^{t-1}}{\vek{r}^t}
$$
The $f^t$'s are again learned linear projections of a pooled bag-of-words embedding, and again we compute the softmax of the appropriate set $\vek{x}^k$ (where $k$ is the number of hops associated with task) and use cross entropy loss.  Results are again shown in Table~\ref{tbl:exp_realqa}.

\textbf{End-to-end QA for very long varying-length reasoning chains.}
To explore performance on even longer reasoning chains, we generated
simple artificial sentences describing long chains of relationships on
a 10-by-10 grid (e.g., ``from center left go down then right'').  We
trained on 360,000 randomly generated sentences containing between 1
and 10 hops, and tested on an addition 1200 sentences.

For vary-length chains, the meaning of each relation set
$\vek{\arel}^t$ can be context-dependent, so than many fixed templates,
we use an encoder-decoder model.  The question is encoded with the
final hidden state of an LSTM, written here $\vek{h}_0$.  We then
generate a reasoning chain of length up to $T$ using a decoder LSTM.
At iteration 0, a distribution over possible starting points (e.g.,
``center left'') is produced, denoted $\vek{x}^0$.  At iteration
$t>0$, the decoder emits a scalar probability of stopping, $p^t$, and
a distribution over relations to follow $\vek{r}^t$, and updates the
model, as we did for the MetaQA dataset, to let \( \vek{x}^t =
\follow{\vek{x}^{t-1}}{\vek{r}^t} \). It also updates the decoder
hidden state to $\vek{h}^t$, as below:  
\begin{equation} \label{eq:vary}
\begin{gathered}
\vek{h}^0 = \textnormal{LSTM}(q);~~~
\vek{x}^0 = f_0(\vek{h}^0);~~~
\vek{r}^0 = \vek{0};~~~
p^0 = 1\\
\textnormal{for $t=1,\ldots,T$:}~~~
p^t = f_p(\vek{h}^{t-1});~~~
\vek{r}^t = f_r(\vek{h}^{t-1});~~~
\vek{h}^t = f_h(\vek{h}^{t-1}, \vek{r}^{t-1}) %\\
%\hat{\vek{a}} = \sum_{t=1}^T \vek{x}^t \cdot p^t \prod_{t'<t} (1-p^{t'})
\end{gathered}
\end{equation}
The final predicted location is a mixture of all the $\vek{x}_t$'s
weighted by the probability of stopping $p_t$ at iteration $t$, i.e.,
\( \hat{\vek{a}} = \textit{softmax}(\sum_{t=1}^T \vek{x}^t \cdot p^t
\prod_{t'<t} (1-p^{t'})) \).  We trained on 360,000 randomly generated
sentences requiring 1 and 10 hops, and tested on an additional 12,000
sentences, with $f_r$ again linear, $f_p$ a logistic function, and
$f_h$ an LSTM cell.  Results are shown in the right-hand side of Figure~\ref{fig:scaling}.

\textbf{Discussion of QA results.} In Table~\ref{tbl:exp_realqa} we
compare our results on the non-synthetic tasks with Key-Value Memory
Network (KV-Mem) baselines \cite{kvmem}.  For the smaller MetaQA
dataset, KV-Mem is initialized with all facts within 3 hops of the
query entities, and for WebQuestionsSP it is initialized by a
random-walk process seeded by the query entities (see
\cite{graftnet,zhang2017variational} for details).  RSF consistently
outperforms the baseline, dramatically so for longer reasoning chains.
The synthetic task shows that there is very little degradation as chain length increases, with Hits@1 for 10 hops still 89.7\%.

We also compare to two much more complex architectures that perform
end-to-end question answering in the same setting used here: VRN
\cite{zhang2017variational} and GRAFT-Net \cite{graftnet}.  Both
systems build question-dependent subgraphs of the KB, and then use
graph CNN-like methods \cite{kipf2016semi} to ``reason'' with these
graphs.  This process is much more complex to perform at test time, so
arguably the systems are not strictly comparable---however, it is
interesting to see that the RSF model is competitive with these
approaches on the most difficult 3-hop setting.

\subsubsection{Knowledge Base Completion Experiments}
\textbf{Knowledge base completion}.  In KB completion, one attempts to recover KB triples that have been removed from a KB by performing inference over the remaining triples.  We looked at two KB completion datasets. The NELL-995 dataset \cite{xiong2017deeppath} has 12 different ``query relations'' (i.e., triples for 12 relations have been removed) paired with a KB with 154k facts, 75k entities, and 200 relations. 
%Similar to MINERVA \cite{das2017go}, we remove all training and testing examples from the graph and discard the test examples whose subject entities are missing from the graph. \hs{this seems a little unclear to me how this can falls into equation \ref{eq:vary}.}

In the KB completion task, a query is a relation name $q$ and a start entity $\vek{x}$, and we assume the answers are computed with the disjunction of multiple inference chains of varying length.  Each inference chain has a maximum length of $T$ and we build $N$ distinct inference chains in total.
\begin{equation*}
\begin{gathered}
\textnormal{for $i=1,\ldots, N$ and $t = 1, \dots, T$:}~~~~
\vek{r}_i^t = f_i^t(q);~~~\vek{x}_i^{t+1} = \follow{\vek{x}_i^t}{\vek{r}_i^t} + \vek{x}_i^t%\\
\end{gathered}
\end{equation*}
The final output is a softmax of the mix of all the $\vek{x}_i^T$'s: i.e., we let
\( \hat{\vek{a}} = \textit{softmax} (\textstyle\sum_{i\in \{1\dots N\}} \vek{x}_i^T) \).
The purpose of the update 
\( \vek{x}_i^{t+1} = \follow{\vek{x}_i^t}{\vek{r}_i^t} + \vek{x}_i^t \) is to give the model access to outputs from chains of length less than $T$. The encoding of $q$ is based on a lookup table, and each relation vector $f_i^t$ is a learned linear transformation of $q$'s embedding.  We tune the hyperparameters $T \in \{1,\ldots, 6\}$ and $N\in\{1,2,3\}$.

% In KB completion task, a query is usually answerable with multiple inference chains and the length of inference chains could vary, so the model should decide the length of paths it follows. In this experiment, the maximum length of an inference chains $M$ and the number of inference chains $N$ are hyper-parameters to be tuned. We design our model to softly decide the length by letting it, at each step, either follow the relation-set $\srel$ on the knowledge base or take a shortcut. For each inference chain, $i \in \{1\dots N\}$, we starts with $\vek{X}_i^0$, the one-hot vector of the subject entity, and run for $M$ iterations. And the final weights are the sum of $\vek{X}_i^M$ across all inference chains:

\textbf{Discussion of KB completion results.} KB completion is often performed with KB embedding methods.  We record several results for two KB embedding baselines and also MINERVA, a state-of-the-art RL-based method.  Baseline results for KB completion are from \cite{das2017go}.
Again the baselines are to some degree incomparable---RL methods are generally more difficult to train, and KB embedding methods do not generalize to entities seen outside of training.  Quantitatively the results are similar to on the QA tasks---the RSF model outperforms the simpler baseline methods, and is competitive with the best existing approaches.

\subsubsection{Discussion of execution time on QA and KB completion} 

We compare the training time of our model with minibatch size of 10 on three different tasks: NELL-995, MetaQA, and WebQuestionsSP. The results are shown in Table \ref{tbl:exe_time} and are consistent with the results of Figure~\ref{fig:scaling}: training is longer with larger KBs, but even with over 40 million facts and nearly 13 million entities from Freebase, it takes less than 10 minutes to run one epoch over WebQuestionsSP (with 3097 training examples).

\begin{table}[t]
\centering
\small
\begin{tabular}{lccc}
\hline
               & NELL-995 & MetaQA  & WebQuestionsSP \\
               \hline
\# Facts        & 154,213 & 196,453 & 43,724,175     \\
\# Entities     & 75,492  & 43,230  & 12,942,798     \\
\# Relations    & 200     & 9       & 616            \\ \hline \hline
Time (seconds) & 44.3    & 72.6    & 1820          \\ \hline
\end{tabular}\\~\\
\caption{Time to run 10K examples for knowledge bases of different size. \label{tbl:exe_time}}
\vspace{-0.25in}
\end{table}

%\vspace{-0.1in}

\section{Related Work}

%\vspace{-0.1in}

The \followop{} operation used here is implemented in an open-source package called NQL, for neural query language. NQL implements a broader range of operations for manipulating KBs, which are described in a short companion paper  \cite{nqlarxivAnon}.  This paper focuses on implementation and evaluation of the \followop{} operation, issues not covered in the companion paper.  

Although NQL is a dataflow language, not a logic, it is semantically related to TensorLog \cite{cohen2017tensorlog}, a probabilistic logic which also can be compiled to Tensorflow. TensorLog is in turn closely related to ``proof-counting'' logics such as stochastic logic programs \cite{DBLP:journals/ml/Cussens01}.
TensorLog and its relatives do not support second-order reasoning, although other non-neural logics in the same line of work are expressive enough to support templates via reification \cite{de2007problog,wang2013programming}.  

In other work, the differentiable theorem prover (DTP) is a differentiable logic which also includes as a template-instantiation approach similar the one described here.  DPT appears to be much less scalable: it has not been applied to KBs larger than a few thousand triples.  The Neural ILP system \cite{yang2017differentiable} uses approaches related to late mixing together with an LSTM controller to perform KB completion and some simple QA tasks, but it is a monolithic architecture focused on rule-learning, while in contrast we propose a re-usable neural component, which can be used in as a component in many different architectures, and a scalable implementation of this.  It is also reported that neural ILP does not scale to the size of the NELL995 task reported here \cite{das2017go}.

Neural architectures like memory networks \cite{weston2014memory}, or other architectures that use attention over some data structure approximating assertions \cite{andreas2016neural,DBLP:journals/corr/abs-1808-09942}
can be used to
build soft versions of \followop{}: however, they also do not scale
well to large KBs, so they are typically only used in cases where a
small amount of information is relevant to a question (e.g.,  \cite{weston2015towards,zhong2017seq2sql}). 

Graph CNNs \cite{kipf2016semi} also can be used for reasoning, and often do use sparse matrix multiplication, but again existing implementations have not been scaled to tens of millions of triples/edges or millions of entities/graph nodes.  Additionally while graph CNNs have been used for reasoning tasks, the formal connection between them and logical reasoning remains unclear.

An alternative to the end-to-end learning approach that is our focus here is reinforcement learning (RL) methods, which have been
used to learn mappings from natural-language questions to
non-differentiable logical representations \cite{liang2016neural,liang2018memory}, one of the applications we consider here.  RL methods have also been applied to KB completion tasks \cite{das2017go,xiong2017deeppath}, another task we explore.  However, the gradient-based approaches enabled by our methods are generally preferred, as being easier to implement and tune on new problems.

%\vspace{-0.1in}

\section{Conclusions}

%\vspace{-0.1in}

In this paper we described scalable differentiable implementations of
the second-order reasoning required to evaluate multi-hop rule
templates.  In particular we introduce a single new operation called
\trm{\followop}, which can be used to compositionally construct
multi-hop templates in a neural model.  The operation is
differentiable, so loss on a proposed template instantiation can be
backpropagated to the functions that instantiate variables in
a template.  

We evaluated a number of alternative implementations of this operation,
which lead to different trade-offs with respect to time and memory,
and show that an appropriate implementation of \followop{} can scale
to KBs with tens of millions of facts, millions of entities, and
thousands of relations on a single modern GPU.  
We also demonstrate experimentally that models based on \followop{}
perform well on a number of tasks that require multi-hop reasoning.
%In particular, the approach allows us to easily learn strategies to
%answer natural language questions on a KB, as well as to learn
%relational concepts using a KB as background knowledge.

The models that are learned have two other advantages, which are not explored in this paper.  They have
interpretable latent variables, corresponding to the template
variables, and on several benchmark tasks the results exceed the
current state-of-the-art.  For the case of question-answering, the
models we propose are compatible with recently-developed approaches
for producing contextual representations of language by large-scale
pre-training \cite{devlin2018bert,peters2017semi}.

The approaches described here suggest a number of further research topics---notably, the question of whether further improvements for QA, KB completion, and other multi-hop reasoning tasks can be obtained by improvements to the very simple architectures that we have used here in combination of \followop{}.  This paper has also focused on $k$-hot encodings of sets, which are suboptimal for small-cardinality sets, suggesting the use of sketching methods for sets, which have been useful in other neural contexts \cite{daniely2016sketching}.  The approach outlined here has also been experimentally applied only to embedding \textit{symbolic} KBs in a neural model, not to embedding ``soft'' KBs, so extending these methods to cover soft KBs would also be of interest.
However, we note that extensions of the set-based \followop{} approach to soft KBs may require additional mechanisms to represent sets of embedded entities \cite{zaheer2017deep,vinyals2015order}.

%\subsubsection*{Acknowledgments}

%\yy{todo}

\bibliographystyle{plain}

\end{document}